\def\year{2021}\relax
\documentclass[letterpaper]{article} 
\usepackage{aaai21}  
\usepackage{times}  
\usepackage{helvet} 
\usepackage{courier}  
\usepackage[hyphens]{url}  
\usepackage{graphicx} 
\usepackage{natbib}  
\usepackage{caption} 
\usepackage{subcaption}
\usepackage{floatrow}
\usepackage{amsmath}
\usepackage{amsthm}
\usepackage{algorithm}
\usepackage{algorithmic}
\usepackage{bm}
\usepackage{amsmath}
\usepackage{amsfonts}
\usepackage{diagbox}
\usepackage{multirow}
\usepackage{booktabs}
\usepackage{algorithm}
\usepackage{algorithmic}
\usepackage{enumerate}
\usepackage{bm}
\usepackage{array}
\usepackage{flushend}

\newfloatcommand{capbtabbox}{table}[][\FBwidth]

\newcommand{\mb}{\mathbb}
\newcommand{\mr}{\mathrm}
\newcommand{\mc}{\mathcal}

\title{Generative Partial Visual-Tactile Fused Object Clustering}
\author{
	Tao Zhang\textsuperscript{\rm 1,2,3},
	Yang Cong\textsuperscript{\rm 1}\thanks{The corresponding author is Prof. Yang Cong and this  work is supported by the National Key Research and Development Program of China (2019YFB1310300) and National Nature Science Foundation of China under Grant (61722311, U1613214, 61821005).},
	Gan Sun\textsuperscript{\rm 1},
	Jiahua Dong\textsuperscript{\rm 1,2,3},
	Yuyang Liu\textsuperscript{\rm 1,2,3},
	Zhenming Ding\textsuperscript{\rm 4} \\
}
\affiliations{
	\textsuperscript{\rm 1}State Key Laboratory of Robotics, Shenyang Institute of Automation, Chinese Academy of Sciences.\\
	\textsuperscript{\rm 2}Institutes for Robotics and Intelligent Manufacturing Chinese Academy of Sciences, Shenyang 110169, China\\
	\textsuperscript{\rm 3}University of Chinese Academy of Sciences,
	\textsuperscript{\rm 4}Department of Computer Science
	Tulane University, USA\\
	zhangtaosia@gmail.com, yangcong81@gmail.com, sungan1412@gmail.com, dongjiahua1995@gmail.com, liuyuyang@sia.cn, zding1@tulane.edu
}

\begin{document}

\maketitle

\begin{abstract}
	Visual-tactile fused sensing for object clustering has achieved significant progresses recently, since the involvement of tactile modality can effectively improve clustering performance.
	However, the missing data (\emph{i.e.,} partial data) issues always happen due to occlusion and noises during the data collecting process.
	This issue is not well solved by most existing partial multi-view clustering methods for the heterogeneous modality challenge.
	Naively employing these methods would  inevitably induce a negative effect and further hurt the performance.
	To solve the mentioned challenges, we propose a \underline{G}enerative \underline{P}artial \underline{V}isual-\underline{T}actile \underline{F}used (\emph{i.e.}, GPVTF) framework for object clustering.
	More specifically, we first do partial visual and tactile features extraction from the partial visual and tactile data, respectively, and  encode the extracted features in modality-specific feature subspaces.
	A conditional cross-modal clustering generative adversarial network is then developed to synthesize one modality conditioning on the other modality, which can compensate missing samples and align the visual and tactile modalities naturally by adversarial learning.
	To the end, two pseudo-label based KL-divergence losses are employed to update the corresponding modality-specific encoders.
	Extensive comparative experiments on three public visual-tactile datasets prove the effectiveness of our method.
\end{abstract}

\section{Introduction}
Benefitting from the great progresses in visual-tactile fused sensing~\cite{liu2018robotic,8460494,DBLP:conf/icra/LeeBL19}, researchers~\cite{zhang2019visual} attempt to focus on visual-tactile fused clustering (VTFC), which aims to group similar objects together in an unsupervised manner.
\begin{figure}[htbp]
	\centering
	\centerline{\includegraphics[width =1.0\columnwidth]{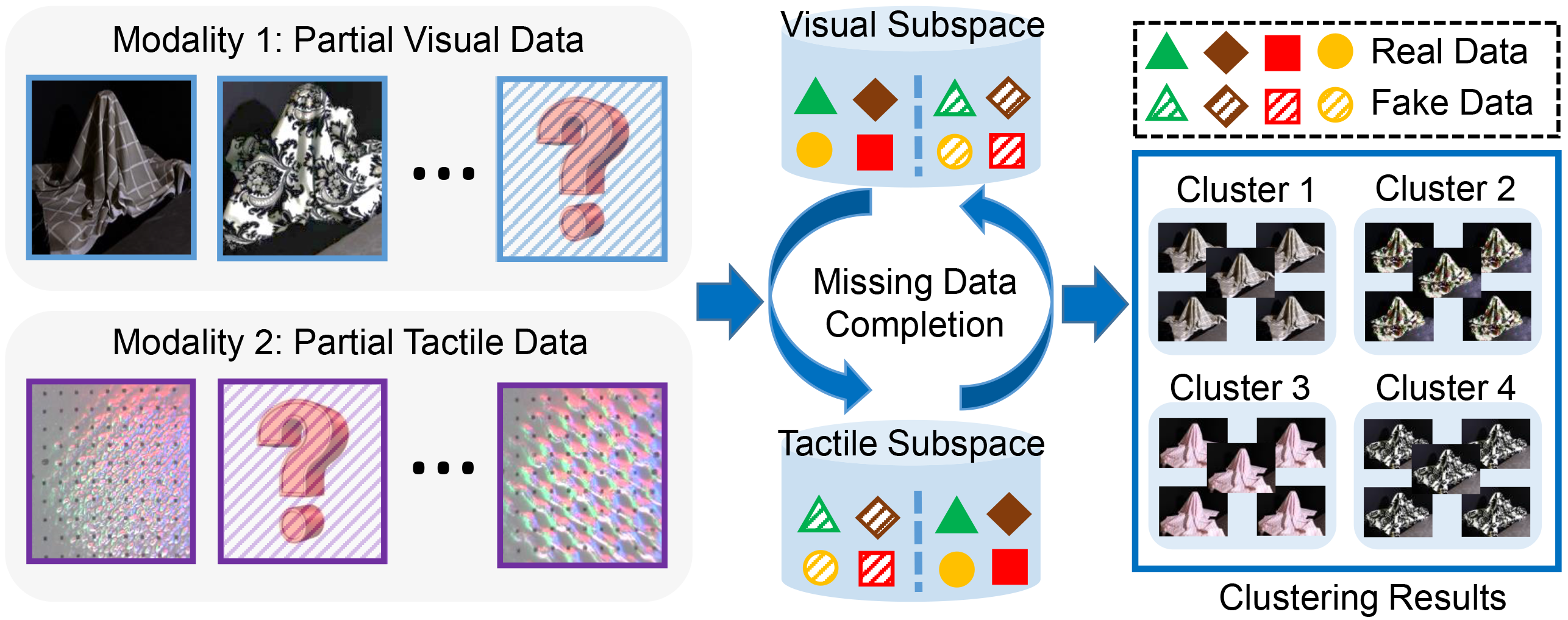}}
	\caption{Diagram  of our proposed method, which first encodes original partial visual and tactile data in modality-specific  subspaces, \emph{i.e.,} visual subspace and tactile subspace. 
		Then, we do visual-tactile fused clustering after completing the missing data. In this way, similar objects are clustered into a group.
	}
	\label{fig:1}
\end{figure}
An interesting example is that when robots employ visual and tactile information to explore unknown environment (e.g., many objects cluttered in an unstructured scene), recognizing the objects in this scene by collecting and annotating a lot of samples is time-consuming and expensive~\cite{Zhaoyangyang2021,wei2019adversarial,ZhaoWYZHW20,ijcai2020-77,sun2020continual}. An alternative solution is to use unsupervised manner to group these objects. In this setting, previous VTFC methods provide a feasible solution by employing fused visual-tactile information in an unsupervised manner to group the objects with same identity into a same group (i.e., object clustering).
Fusion visual-tactile information could improve the clustering performance effectively, since they can provide complementary information.
Generally, most existing VTFC methods mainly utilize the idea of multi-view clustering~\cite{dang2020multi,hushizheDMIB,PR2020DAMC}, \emph{e.g.}, Zhang et al.~\cite{zhang2019visual} propose a VTFC model based on non-negative matrix factorization (NMF) as well as consensus clustering and achieve great progresses. As far as we know, this is the first work about visual-tactile fused clustering.

However, the task of VTFC has not been well settled due to the following challenges \emph{i.e.}, \textbf{partial data} and \textbf{heterogeneous modality}.
\textbf{Partial data:} Existing visual-tactile fused object clustering methods~\cite{zhang2019visual} make a strong assumption that all the visual-tactile modalities well aligned and complete.
However, visual-tactile data usually tend to be incomplete in real world applications.
For instance, when a robot grasps an apple, the visual information of the apple becomes unobservable due to the occlusion of a robot hand.
Moreover, noises, signal loss and malfunction in the data collecting process might make the instance missing.
For instance, in special situations (\emph{e.g.,} underwater scenes), the visual can be easily missing due to turbidity of the water.
These cases mentioned above lead to the incompleteness of multi-modality data, which further hurt the clustering performance.
\textbf{Heterogeneous modality:} Most previous partial multi-view clustering methods use different feature description methods (\emph{e.g.}, SIFT, LBP, HOG) to extract different view features for visual data, which are essentially  homogeneous data.
Therefore, directly employing these methods on heterogeneous data (\emph{i.e.}, visual and tactile data) could induce a negative effect and even unsuccessful clustering task, since they ignore the distinct properties between visual and tactile modalities.

To solve these problems mentioned above, as shown in Figure~\ref{fig:1}, we propose a Generative Partial Visual-Tactile Fused (\emph{i.e.}, GPVTF) framework for object clustering, which aims to obtain better clustering results by adopting generative adversarial learning as well as simple yet effective KL-divergence losses.
Specifically, we first extract partial visual and tactile features from the raw input data, and employ two modality-specific encoders to project the extracted features into visual subspace and tactile subspace, respectively.
Then visual (or tactile) conditional cross-modal clustering generative networks are trained to reproduce tactile (or visual) latent representations in the modality-specific subspaces.
In this way, the our proposed approach is able to effectively leverage the complementary information, and learns the latent subspace level pairwise cross-modal knowledge among visual-tactile data.
The conditional clustering generative adversarial networks can not only complete the missing data, but also force the heterogeneous modalities to be similar and further align them.
With the well completed and aligned  visual and tactile  subspaces, we can obtain expressive representations of the raw visual-tactile data.
Moreover, two pseudo label based fusion KL-divergence losses are employed to update the encoders, and further help 
obtaining better representations for better clustering performance.
Finally, extensive experimental results on three real-world visual-tactile datasets prove the superiority of our proposed framework.
We summarize the contributions of our work as follows:
\begin{itemize}
	\item We put forward a Generative Partial Visual-Tactile Fused (GPVTF) framework for partial visual-tactile clustering.
	To our best knowledge, this is an earlier work about visual-tactile fused clustering, which tackles the problem of incomplete data.
	\item A conditional cross-modal clustering generative adversarial learning schema is encapsulated in our model to complete the missing data and align visual-tactile data, which can further help explore the shared complementary information among multi-modality data.
	\item We conduct comparisons and experiments with three benchmark real-world visual-tactile datasets, which show the superiority of the proposed GPVTF framework.
\end{itemize}

\section{Related Work}
\subsection{Visual-Tactile Fused Sensing}
Significant progresses have been made on visual-tactile fused sensing~\cite{liu2018robotic} in recent years, \emph{e.g.}, object recognition, cross-modal matching and object clustering.
For example, Liu et al.~\cite{liu2016visual} develop an effective fusion strategy for weakly paired visual-tactile data based on joint sparse coding, which makes great success in household object recognition.
Wang et al.~\cite{wang20183d} predict the shape prior of an object from a single color image and then achieve accurate 3D object shape perception by actively touching the object.
Yuan et al.~\cite{yuan2017connecting} show that there is an intrinsic connection between visual and tactile modalities through the physical properties of materials.
Li et al.~\cite{li2019connecting} uses a conditional generative adversarial network to generate pseudo visual (or tactile) outputs based on tactile (or visual) inputs, then expanding the generated data to classification tasks.
Zhang et al.~\cite{zhang2019visual} first propose a visual-tactile fusion object clustering framework base on non-negative matrix factorization (NMF).
However, all of the methods assume that data are  well aligned and complete, which is unrealistic in practical applications.
Thus, we design the GPVTF framework to address these problems for object clustering in this paper.
\begin{figure*}[htbp]
	\centering
	\centerline{\includegraphics[width=1.0\columnwidth]{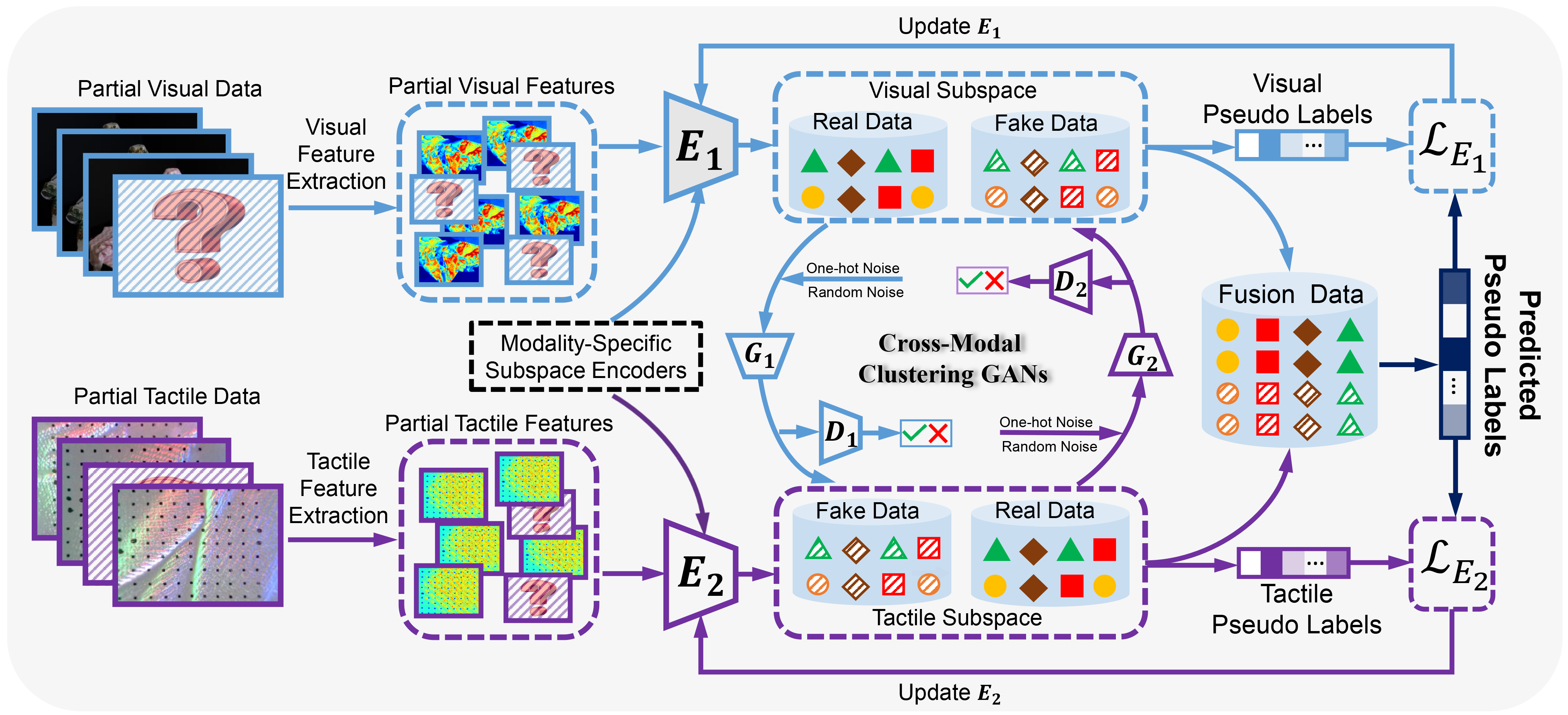}}
	\caption{
		Illustration of the proposed generative partial visual-tactile fused object clustering framework.
		Firstly, partial visual and tactile features are extracted from the raw partial visual and tactile data.
		Then two modality-specific encoders, \emph{i.e.}, visual encoder $E_1(\cdot)$ and tactile encoder $E_2(\cdot)$ are introduced to obtain distinctive representations in visual subspace and tactile subspace. 
		Two cross-modal clustering generators $G_1(\cdot)$ and $G_2(\cdot)$ generate representations conditionally based on the other subspace, which could not only pull the distance between the visual and tactile subspaces but also complete the missing items.
		Finally, both real and generated fake representations are fused to predict the clustering labels.
		Meanwhile, the modality-specific encoders are updated by the KL-divergence losses $\mc{L}_{E_1}$ and $\mc{L}_{E_2}$, which are calculated by the predicted pseudo labels.
	}
	\label{fig:2}
\end{figure*}

\subsection{Partial Multi-View Clustering}
Partial multi-view clustering~\cite{SunCWLF20,li2014partial,Wang2020icmsc,wang2018partial}, which provides a framework to solve the issue of incomplete (partial) input data, can be divided into two categories.
The first category is based on traditional technique, such as NMF and kernel learning.
For example, Li et al.~\cite{li2014partial} propose a incomplete multi-view clustering framework by establishing a latent subspace based on NMF, where incomplete multi-view information is maximized.
Shao et al.~\cite{shao2013clustering} propose a collective kernel learning method to complete missing data and then do clustering tasks.
The second category utilizes generative adversarial networks (GANs) to complete the missing data, for the reason that GANs can align heterogeneous data and complete partial data~\cite{What_Transferred_Dong_CVPR2020,Semantic_Transferable_Dong_ICCV2019,yang2020adversarial,JiangXYCH19}.
For instance, Xu et al.~\cite{xu2019adversarial} propose an adversarial incomplete multi-view clustering method, which performs missing data inference via GANs and learns the common latent subspace of multi-view data simultaneously.
All the methods mentioned above are developed for homogeneous data, they ignore the huge gap between heterogeneous data (\emph{i.e.}, visual and tactile data).

\section{The Proposed Method}
In this section, the proposed Generative Partial Visual-Tactile Fused (GPVTF) framework is presented in detail, together with its implementation.

\subsection{Details of the Model Pipeline}
Given the visual-tactile data $V$ and $T$, where $V$ denotes the visual data (\emph{i.e.}, RGB images) and $T$ denotes the tactile data.
Noticing that the visual and tactile data collected from different tactile sensors lie in different data spaces.
Our proposed GPVTF model consists of two partial  feature extraction processes, \emph{i.e.,} visual feature extraction and tactile feature extraction, which learn partial visual features $X_n^{(1)} \in \mb{R}^{d_1 \times n}  $ from $V$ and tactile features $X_n^{(2)} \in \mb{R}^{d_2 \times n} $ from $T$, where $d_1$ and $d_2$ are the feature dimensions and $n$ is the number of samples;
two modality-specific encoders, ${E_1(\cdot)}$ and ${E_2(\cdot)}$;
two generators, $G_1(\cdot)$ and $G_2(\cdot)$ and their corresponding discriminators, $D_1(\cdot)$ and $D_2(\cdot)$;
two KL-divergence based losses, as illustrated in Figure~\ref{fig:2}.
More details are provided in the following sections.
Particularly, since each dataset has different feature extraction processes, the details of these processes are given in the ``Experiments" section.

\textbf{Encoders and Clustering Module:} Modality-specific encoders $E_1(\cdot)$ and $E_2(\cdot)$ are introduced to project both partial visual and tactile features into the modality-specific subspaces,  \emph{i.e.,} visual subspace and tactile subspace, respectively.
Specifically, in the modality-specific subspaces, the learn latent subspace representations are learned via $Z_{n}^{(m)} = E_m(X_{n}^{(m)};{\theta}_{E_m})$, where $m=1$ denotes the visual modality, $m=2$ denotes the tactile modality, and ${\theta}_{E_m}$ denote the network parameters of the $m$-th encoder.
Then the fused representations (\emph{i.e.}, $m=3$ ) can be gained by:
\begin{equation}\label{eq:Fusion}
\begin{aligned}
Z_n^{(3)} = (1-\alpha)Z_n^{(1)} +  \alpha Z_n^{(2)},
\end{aligned}
\end{equation}
where $\alpha > 0 $ is the weighting coefficient that balances the ratio of tactile and visual modalities.
Next, the K-means method is employed on $Z^{(m)}_n$ to get the initial clustering centers $\{ {\mu}^{(m)}_j \}_{j=1}^k$, where $k$ is the number of clusters\footnote{Since we do clustering according to object identity, the $k$ is set to be equal with the number of types of objects in the datasets. 
	Specifically, $k$ is set to be $53$, $119$ and $108$ for \textbf{PHAC-2}, \textbf{GelFabric}, and \textbf{LMT} datasets, respectively.}.
Inspired by~\cite{xie2016unsupervised}, we employ Student's t-distribution to measure the similarity of latent subspace representations $Z^{(m)}_n$ and the clustering center ${\mu}^{(m)}_j$:
\begin{equation}\label{eq:Q}
\begin{aligned}
q^{(m)}_{nj} = \frac{ {(1+\|Z^{(m)}_n-\mu^{(m)}_j\|^2/\gamma)}^{-\frac{2}{\gamma+1}}} { \sum_{j'}(1+\|Z^{(m)}_n-\mu_{j'}^{(m)}\|^2 / \gamma)^{- \frac{\gamma+1}{2}}},
\end{aligned}
\end{equation}
where $\gamma$ is the degrees of freedom of the Student's t-distribution and set to be $1$ in this paper; $q^{(m)}_{nj}$ are the pseudo-labels, which denote the probability of assigning sample $n$ to cluster $j$ for the $m$-th modality.

To improve cluster compactness, we pay more attention to data points of which are assigned with high confidence, by obtaining the target distribution $p_{nj}^{(m)}$ as follows:
\begin{equation}\label{eq:P}
\begin{aligned}
p^{(m)}_{nj} = \frac{{q^{(m)}_{nj}}^2 \big/ \sum_nq^{(m)}_{nj}}{\sum_{j'}{q^{(m)}_{nj'}}^2
	\big/\sum_nq^{(m)}_{nj}}.
\end{aligned}
\end{equation}
Then the encoders are trained with fused KL-divergence losses, which are defined as follows:
\begin{equation}\label{eq:E1_KL}
\begin{aligned}
\mc{L}_{E_m}& \!=\!KL \big(P^{(m)}\vert\vert Q^{(m)}\big) + \beta KL \big(P^{(3)}\vert\vert Q^{(3)}\big) \\
&  \!=\! \sum_{n}\sum_{j}p^{(m)}_{nj}\log\frac{p^{(m)}_{nj}}{q^{(m)}_{nj}} \!+\! \beta \sum_{n}\sum_{j}p^{(3)}_{nj}\log\frac{p^{(3)}_{nj}}{q^{(3)}_{nj}},
\end{aligned}
\end{equation}
where $m =1$ and $m =2$ correspond to the losses of encoders $E_1(\cdot)$ and $E_2(\cdot)$, and $\beta$ is a trade-off parameter.
The encoders are implemented by a two-layer fully-connected network.

\textbf{Conditional Cross-Modal Clustering GANs:} Noticing that the gap between visual and tactile modalities is very large since their frequency, format and receptive field are quite different.
Thus, directly employing GANs in the original space $X_n^{(m)}$ might increase the difficulty of training or even lead to non-convergence.
To address this challenge, we develop a conditional cross-modal clustering GANs, which generates one latent space conditional on the other latent space.
Specifically, the conditional cross-modal cluster GANs including $G_m(\cdot)$ and $D_m(\cdot)$, where $G_m(\cdot)$ competes with $D_m(\cdot)$ to generate  samples as real as possible, and the loss function is given as:
\begin{equation}\label{eq:G1d}
\begin{aligned}
\mc{L}_{G_{md}}= -E_{\omega \sim P_\omega(\omega)}\log\big( 1 \!-\! D_m (G_m(\omega|Z_n^{(m)})) \big),
\end{aligned}
\end{equation}
where $\omega$ is the noise matrix.
Noticing that our goal is clustering rather than generation, a prior that consists of normal random variables cascaded with one-hot noise is sampled, which is different from tradition GANs.
More specifically, $\omega = (\omega_n,\omega_c)$, $\omega_n \sim N(0,\sigma^2I_{dn})$, $\omega_c = e_k$, $e_k$ is the $k$-th elementary vector in $\mb{R}^k$ and $k$ is the number of clusters. 
We choose $\sigma = 0.1$ in all our experiments.
By this way, a non-smooth geometry latent subspace is created, and $G_m(\cdot)$ can generate more distinctive and robust representations which are beneficial for clustering performance, \emph{i.e.}, not only the gap between visual and tactile modalities can be mitigated but also the missing data are completed naturally.

Moreover, since training the GANs in Eq.~\eqref{eq:G1d} is not trivial~\cite{wang2019generative}, a regularizer, which forces  the real samples and the generated fake samples to be similar, is introduced to obtain stable generative results, which can be defined as:
\begin{equation}\label{eq:G1s}
\begin{aligned}
\mc{L}_{G_{ms}} = E_{\omega \sim P_\omega(\omega)} \big( \|G_m(\omega|Z_n^{(m)})) - Z_n^{(m)}\|^2 \big).
\end{aligned}
\end{equation}
Then, the overall loss function of $G_m(\cdot)$ is given as follows:
\begin{equation}\label{eq:G}
\begin{aligned}
\mc{L}_{G_m} = \mc{L}_{G_{md}} + \lambda \mc{L}_{G_{ms}},
\end{aligned}
\end{equation}
where $\lambda$ is a trade-off parameter which balances the two losses and is set to be 0.1 in this paper.
$G_m(\cdot)$ is a three-layer network.

The discriminator $D_m(\cdot)$ is designed to discriminate the fake representations generated by $G_m(\cdot)$ and  the real representations in the modality-specific subspaces.
The object function for $D_m(\cdot)$ can be given as:
\begin{equation}\label{eq:D1}
\begin{aligned}
\mc{L}_{D_m}= &E_{Z \sim P_Z(Z)}\log D_m( E_m(X_n^{(m)};\theta_{E_m})) +  \\
& E_{\omega \sim P_\omega(\omega)}\log \big(1 \!-\! D_m( G_m(\omega| E_m(X_n^{(m)};\theta_{E_m})) ) ) \big).
\end{aligned}
\end{equation}
The proposed $D_m(\cdot)$ is mainly made up of a fully connected layer with ReLU activation, a mini-batch layer~\cite{salimans2016improved} that can increase the diversity of fake representations, a sigmoid function which outputs the fake-real possibility of input representations.
Then, both the generated fake and real representations are fused. 
Thus, the fusion representations Eq.~\eqref{eq:Fusion} can be modified to:
\begin{equation}\label{eq:Fusion_fake}
\begin{aligned}
Z^{(3)}_n =  (1-\alpha)Z_n^{(1)} + \alpha  Z_n^{(2)} + \sum_{m=1}^{2}\varphi_{m}Z_{\mr{fake}}^{(m)},
\end{aligned}
\end{equation}
where $\varphi_{m}$ is the weighting coefficients of real and the generated fake representations for the $m$-th modality, \emph{i.e.,} $m=1$ represents visual modality and $m=2$ represents tactile modalities, respectively. 
The overall loss function of our model is summarized as follows:
\begin{equation}\label{eq:Total_Loss}
\begin{aligned}
\mc{L}_{total} = \min_{E_{m},G_m}\max_{D_m}\mc{L}_{E_{m}} + \mc{L}_{G_{m}}+ \mc{L}_{D_{m}},
\end{aligned}
\end{equation}
where $\mc{L}_{E_{m}}$ are the KL-divergence losses, $\mc{L}_{G_{m}}$ and $\mc{L}_{D_{m}}$ are the conditional cross-modal clustering GANs losses.
\begin{algorithm}[htbp]
	\caption{Training Process of the Proposed Framework}
	\label{alg:updateALL}
	\begin{algorithmic}[1] %
		\STATE  \textbf{Input:}Visual-tactile data:\{$V$, $T$\}. Number of clusters: $k$. The maximum number of iterations: MaxIter; hyper-parameters $\alpha$, $\beta, \varphi_1$ and $\varphi_2$.
		\STATE  \textbf{Initialization:} Project \{$V$, $T$\} into feature subspaces \{$X_n^{(1)}$, $X_n^{(2)}$\}.Initialize the parameters of the networks with Xavier initializer. Calculate the initial fusion representations $Z_n^{(3)}$ and the clustering centers $\{\mu_j^{(m)}\}_{j=1}^k$.
		\FOR{iter $\leq$ MaxIter}
		\STATE Train the encoders $E_m{(\cdot)}$ with corresponding KL-divergence losses $\mc{L}_{E_m}$, $\forall m = 1,2$.
		\STATE Train the generators $G_m(\cdot)$ with  $\mc{L}_{G_m}$, $\forall m = 1,2$.
		\STATE Train the discriminators $D_m(\cdot)$ with $\mc{L}_{D_m}$, $\forall m = 1,2$.
		\STATE Update the fused representation $Z_n^{(3)}$ and clustering centers $\{\mu_j^{(m)}\}_{j=1}^k$, $\forall m = 1,2$.
		\ENDFOR
		\STATE Gain the updated fusion representation $Z_n^{(3)}$, fusion clustering centers $\{\mu_j^{(3)}\}_{j=1}^k$ and pseudo-labels $q_{nj}^{(3)}$.
		\STATE Predict the clustering labels according to $q_{nj}^{(3)}$.
		\RETURN Predicted cluster labels.
	\end{algorithmic}
\end{algorithm}

\subsection{Training}
The whole process of the proposed GPVTF framework is summarized as below.

\emph{\textbf{Step 1} Initialization}: We feed the partial visual and tactile features $X_n^{(1)}$ and $X_n^{(2)}$ into $E_1(\cdot)$ and $E_2(\cdot)$ to obtain the initial latent subspace representations $Z_n^{(m)}$.
Then standard K-means method is applied on  $Z_n^{(m)}$ to get the initial clustering centers $\{\mu_j^{(m)}\}_{j=1}^k, \forall m = 1,2,3$.

\emph{\textbf{Step 2} Training encoders}: Eq.~\eqref{eq:Q} is employed to calculate the pseudo-labels $q_{nj}^{(m)}$;
$p_{nj}^{(m)}$ and KL-divergence losses $L_{E_m}$ are computed by Eq.~\eqref{eq:P} and Eq.~\eqref{eq:E1_KL}, respectively.
Then $L_{E_m}$ are fed to its corresponding Adam optimizers to train the encoders and the learning rates are set to be 0.0001.

\emph{\textbf{Step 3} Training conditional cross-modal clustering  GANs}: In this step, we employ the generator losses, \emph{i.e.}, Eq.~\eqref{eq:G1d} and Eq.~\eqref{eq:G1s} with Adam optimizers to update the  parameters of the two generators and the learning rates are set to be 0.000003 and 0.000004 for $G_1(\cdot)$ and $G_2(\cdot)$, respectively.
Next, the two discriminators $D_1(\cdot)$ and $D_2(\cdot)$ are optimized by Eq.~\eqref{eq:D1} with Adam optimizers and the leaning rates are set to be 0.000001 both for $D_1(\cdot)$ and $D_2{(\cdot)}$.
We update the generators five times while updating the discriminators once.

\emph{\textbf{Step 4}} After the framework is optimized, we feed original data to the model and then obtain the completed fusion representations $Z_n^{(3)}$ as well as the updated clustering centers $\{\mu_j^{(m)}\}_{j=1}^k$.
Then the predicted clustering labels $q^{(3)}_{nj}$ are calculated by Eq.~\eqref{eq:Q}.
Finally, we choose the maximum value of $q^{(3)}_{nj}$ as the predicted clustering labels.
We implement the model with Tensorflow 1.12.0, and set
the batch size to be 64.
We summarize the overall training process of the proposed framework in \textbf{Algorithm~\ref{alg:updateALL}}.

\section{Experiments}
In this section, the used datasets, comparison methods, evaluation metrics and  experimental results are given. 
\subsection{Datasets and Partial Data Generation}
\begin{figure}[htbp]
	\centering
	\centerline{\includegraphics[width =1.0\columnwidth]{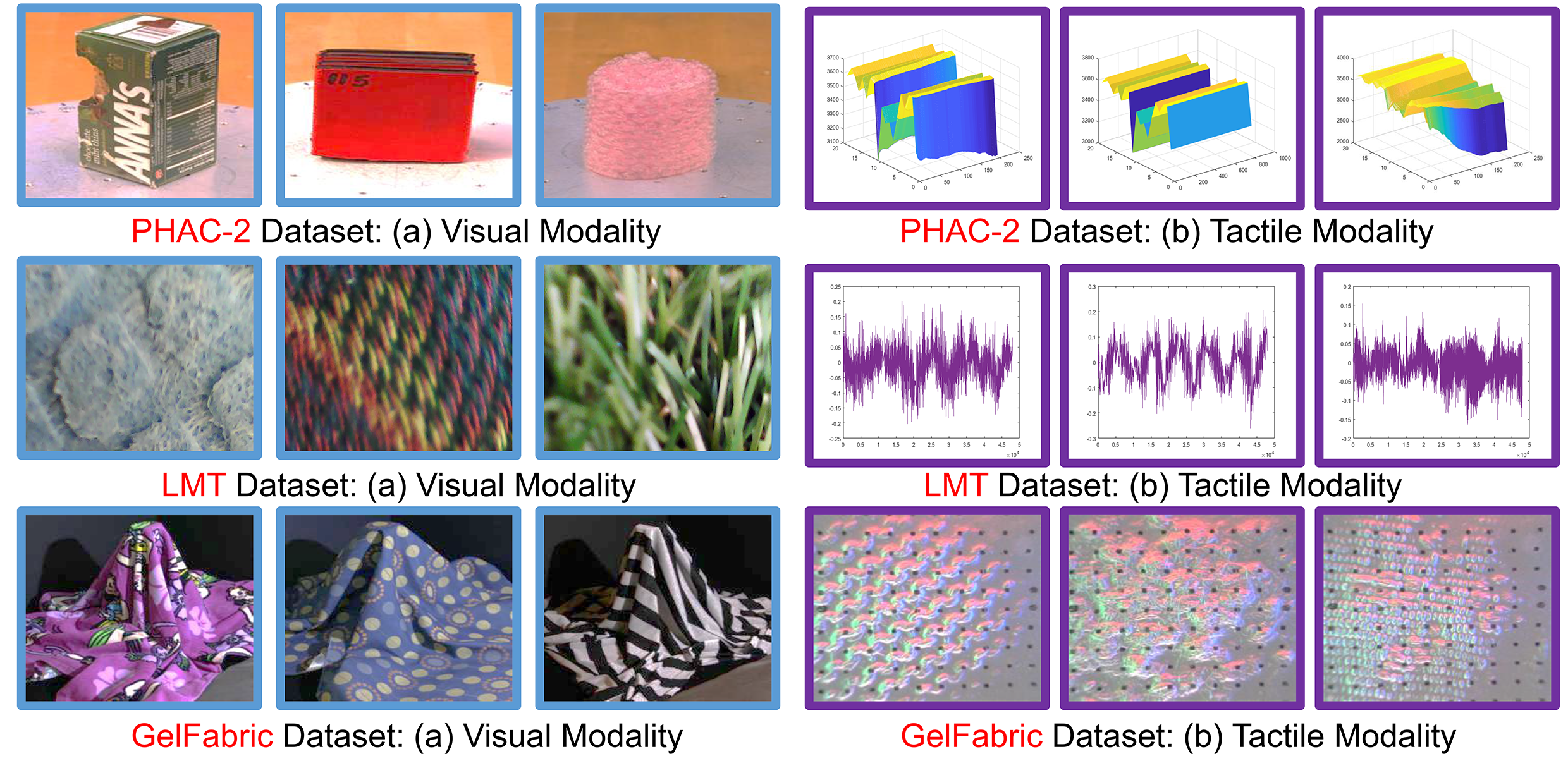}}
	\caption{Example visual images and tactile data of the used datasets, \emph{i.e.,} GelFabric dataset~\cite{yuan2017connecting}, LMT dataset~\cite{strese2015surface,Strese2014A}, and PHAC-2 dataset~\cite{gao2016deep}.
		Intuitively, there are intrinsic differences between visual and tactile data.
		It is worth noting that the tactile signals in the PHAC-2 dataset consist of multiple components.
		We only visualize the electrode impedance component for simplicity.
		More details of these datasets can be found in their corresponding references.
	}
	\label{fig:3}
\end{figure}

\textbf{PHAC-2}~\cite{gao2016deep} dataset consists of color images and tactile signals of 53 household objects, where each object has 8 color images and 10 tactile signals. 
We use all the images and the first 8 tactile signals  to build the initial paired visual-tactile dataset in this paper. 
The feature extraction process of the tactile modality is similar with~\cite{gao2016deep,zhang2019visual}, and the visual features are extracted by  AlexNet~\cite{krizhevsky2012imagenet}, which is pre-trained on the ImageNet. 
After feature extraction, 
4096-D visual and 2048-D tactile features are obtained. 
\textbf{LMT}~\cite{zheng2016deep,strese2015surface} dataset consists of 10 color images and 30 haptic acceleration data of 108 different surface materials. 
The first 10 haptic acceleration data and all the images are used. 
We extract 1024-D tactile features similarly with~\cite{liu2019lifelong} and 4096-D visual features by the pre-trained AlexNet.
\textbf{GelFabric}~\cite{yuan2017connecting} dataset includes visual data (\emph{i.e.,} color and depth images) and tactile data of 119 kind of different fabrics.
Each fabric has 10 color images and 10 tactile images, which are used in this paper. 
Since both the visual and tactile data are image formats, we extract 4096-D visual and tactile features with pre-trained AlexNet.
Some examples of the used datasets are given in Figure~\ref{fig:3}.

\textbf{Partial data generation}: The partial visual-tactile datasets are generated in a similar way with partial multi-view clustering settings,
\emph{e.g.}, Xu et al~\cite{xu2019adversarial}.
Supposing that the number of all the visual and tactile samples is $N$ in each dataset, we randomly select $\tilde{N}$ samples as the missing data points.
Then, the Missing Rate (\emph{i.e.}, $\mc{MR}$) can be defined as $\mc{MR} = \frac{\tilde{N}}{N}$.

\begin{table*}[htbp]
	\centering
	\caption{ACC and NMI performance  on the three visual-tactile datasets, when the missing rate is set to be 0.1.}
	\begin{tabular}{|ccc|cc|cc|}
		\hline
		\multicolumn{3}{|c|}{ PHAC-2 Dataset} & \multicolumn{2}{c|}{ LMT Dataset}
		& \multicolumn{2}{c|}{GelFabric Dataset} \\
		\hline
		Method & ACC($\%$) & NMI($\%$) 
		& ACC($\%$) & NMI($\%$)      
		& ACC($\%$) & NMI($\%$)  \\
		\hline\hline
		SC1& 40.62$\pm$0.64 & 67.05$\pm$0.60
		&51.32$\pm$1.19 & 76.07$\pm$0.32
		&49.50$\pm$0.69 & 72.98$\pm$0.31 \\
		
		SC2& 30.20$\pm$0.95  &  56.67$\pm$0.60 
		& 15.02$\pm$0.26  &  42.61$\pm$0.27 
		& 45.87$\pm$0.76  &  72.92$\pm$0.34 \\
		
		ConcatPCA& 45.38$\pm$1.04  &  69.17$\pm$0.64 
		& 40.78$\pm$0.48  &  68.16$\pm$0.21 
		& 47.95$\pm$1.64  &  74.56$\pm$0.84 \\
		
		GLMSC& 37.38$\pm$0.17  &  64.57$\pm$0.47 
		& 41.30$\pm$1.11  &  68.37$\pm$0.83
		& 50.88$\pm$1.01  &  75.55$\pm$0.14 \\
		
		VTFC& 51.41$\pm$0.63   &  70.85$\pm$0.32   
		& 43.94$\pm$0.16   &  51.03$\pm$0.22    
		& 55.72$\pm$1.04   &  74.76$\pm$0.38 \\
		
		IMG& 37.90$\pm$0.92    &  49.79$\pm$0.14    
		& 41.66$\pm$1.68    &  67.45$\pm$0.93
		& 37.39$\pm$2.10    &  66.06$\pm$0.48 \\
		
		GRMF& 33.16$\pm$1.62  &  60.54$\pm$0.73  
		& 26.59$\pm$0.71  &  57.89$\pm$0.37 
		& 40.97$\pm$0.99  &  72.69$\pm$0.37  \\
		
		UEAF& 40.56$\pm$0.06  &  63.20$\pm$0.39
		& 47.78$\pm$0.19  &  74.09$\pm$0.60
		& 51.26$\pm$0.05  &  72.36$\pm$0.72  \\
		
		\textbf{OURS}& \textbf{53.30$\pm$0.69}  & \textbf{74.47$\pm$0.18}
		& \textbf{54.81$\pm$1.36}  & \textbf{80.37$\pm$0.40}
		& \textbf{59.89$\pm$0.42}  & \textbf{81.60$\pm$0.37}\\
		\hline
	\end{tabular}%
	\label{tab:tb1}%
\end{table*}%

\subsection{Comparsion Methods and Evaluation Metrics}
We compare our GPVTF model with the following baseline methods.
We first employ standard spectral clustering methods on the modality-specific features, \emph{i.e.}, visual features $X_n^{(1)}$ and tactile features $X_n^{(2)}$, which are termed as \textbf{SC1} and \textbf{SC2}.
\textbf{ConcatPCA} concatenates feature vectors of different modalities via PCA and then performs standard spectral clustering.
\textbf{GLMSC}~\cite{zhang2018generalized} proposes a subspace multi-view clustering model under the assumption that each single feature view originates from one comprehensive latent representations.
\textbf{VTFC}~\cite{zhang2019visual} is a pioneering work to incorporate visual modality with tactile modality in the object clustering tasks based on auto-encoders and NMF.
\textbf{IMG}~\cite{zhao2016incomplete} does the incomplete multi-view clustering by transforming the original partial data to complete representations.
\textbf{GRMF}~\cite{wen2018incomplete} exploits the complementary and local information among all views and samples based on graph regularized matrix factorization.
\textbf{UEAF}~\cite{wen2019unified} performs missing data inference with locality-preserved constraint.

\textbf{Evaluation Metrics:} Two widely used clustering evaluation metrics, \emph{i.e.}, Accuracy (ACC) and Normalized Mutual Information (NMI) are employed to assess the effectiveness of the clustering performance.
For all the metrics, higher value indicates better performance.
More details of these metrics can be found in~\cite{schutze2008introduction}.

\begin{figure*}[!t]
	\centering
	\centerline{\includegraphics[width =1.0\columnwidth]{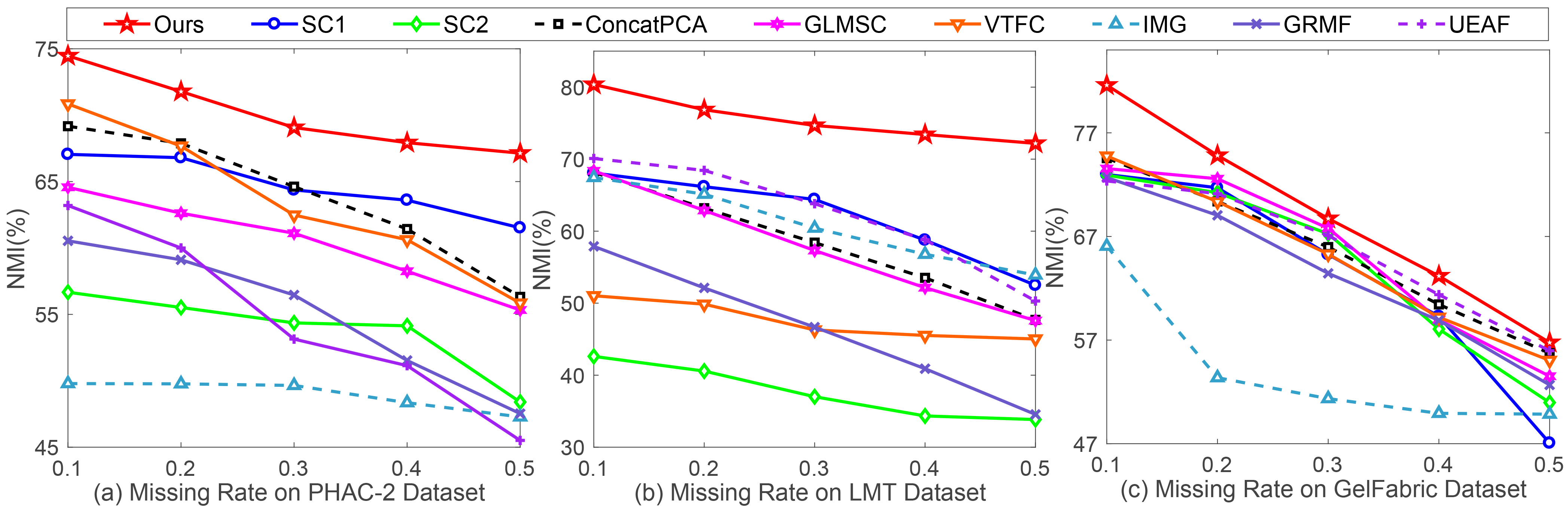}}
	\caption{The average clustering NMI performance with respect to different missing rate on the (a) PHAC-2 dataset, (b) LMT dataset and (c) GelFabric dataset.
	}
	\label{fig:4}
\end{figure*}

\begin{figure*}[!t]
	\centering
	\centerline{\includegraphics[width =1.0\columnwidth]{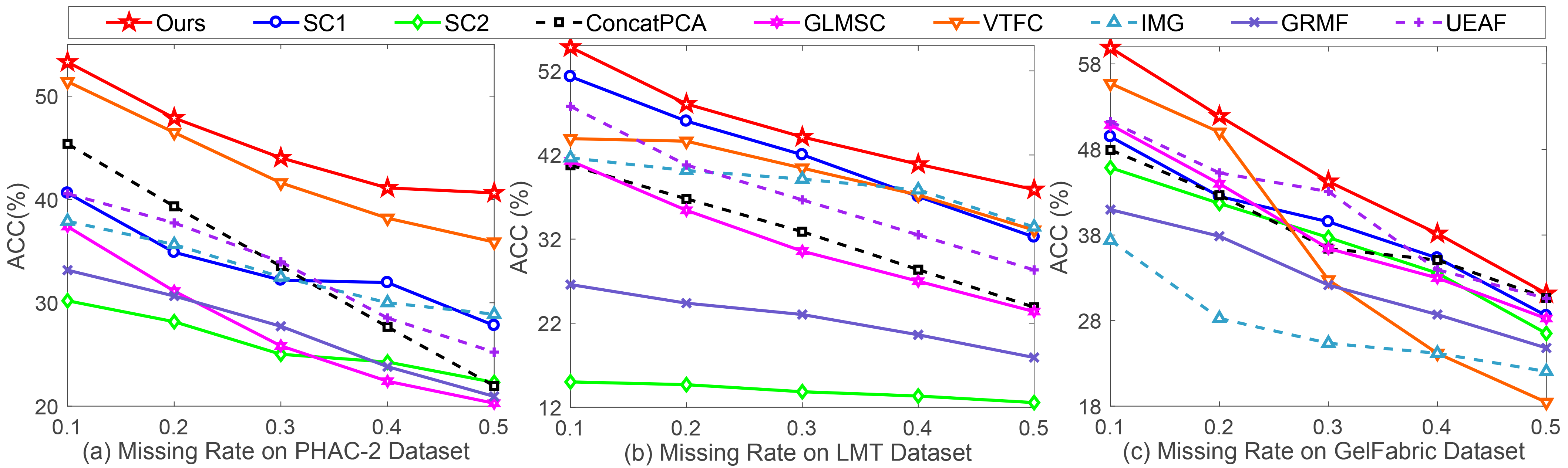}}
	\caption{The average clustering ACC performance with respect to different missing rate on the (a) PHAC-2 dataset, (b) LMT dataset and (c) GelFabric dataset.
	}			
	\label{fig:5}
\end{figure*}

\subsection{Experimental Results}
Experimental results on three public visual-tactile datasets are reported by comparing with the state-of-the-arts in this subsection.
Due to the randomness of missing data generation, all experiments are repeated in ten times and reported with the mean value.
Generally, the observations are summarized as follows:
\textbf{1}) As shown in Table~\ref{tab:tb1}, where the missing rate is set to be 0.1, our GPVTF
model consistently outperforms other methods with a clear improvement.
\begin{figure}[htbp]
	\centering
	\subcaptionbox{ACC ($\%$) performance}{\includegraphics[width =.49\columnwidth]
		{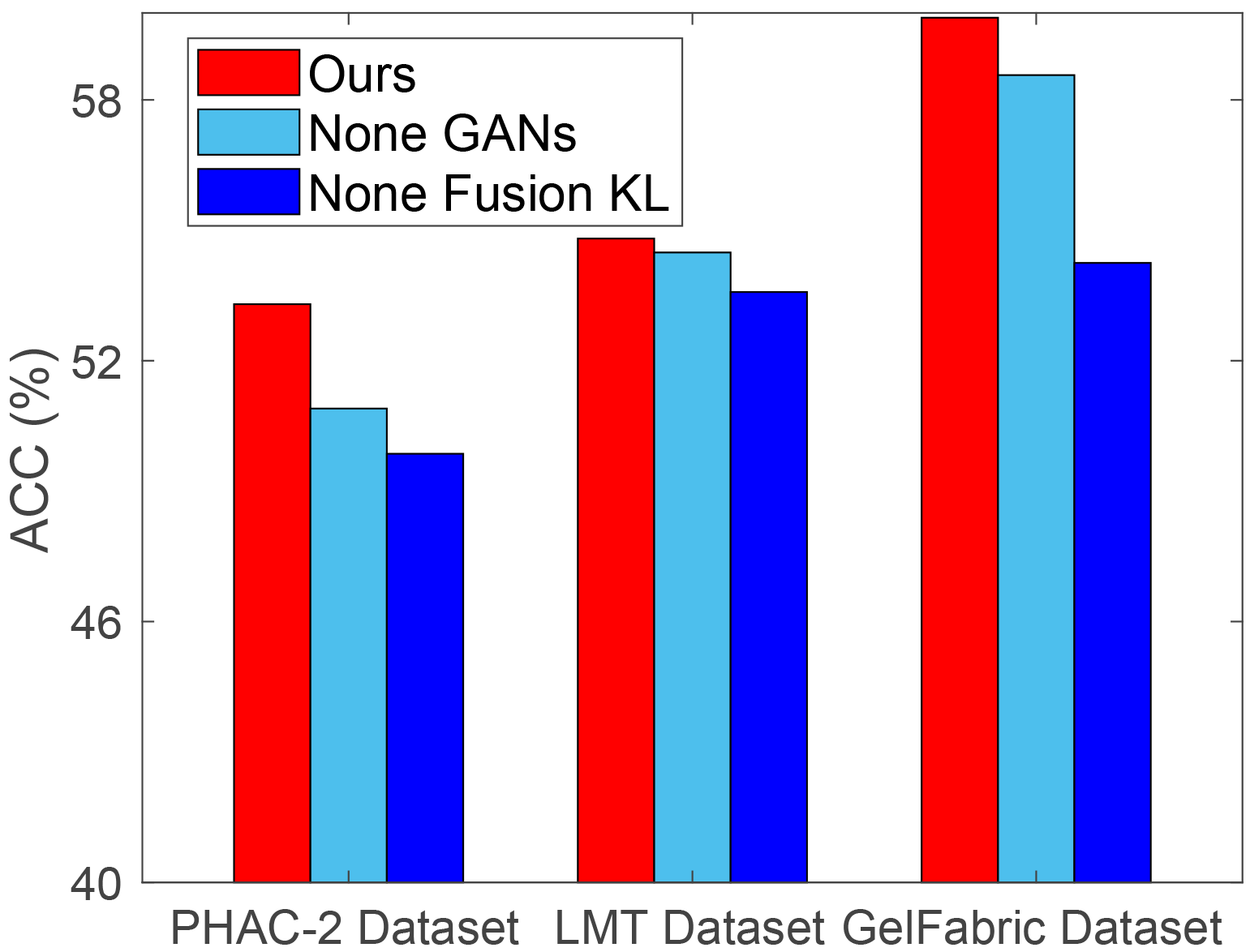}}
	\subcaptionbox{NMI ($\%$) performance}{\includegraphics[width =.49\columnwidth]
		{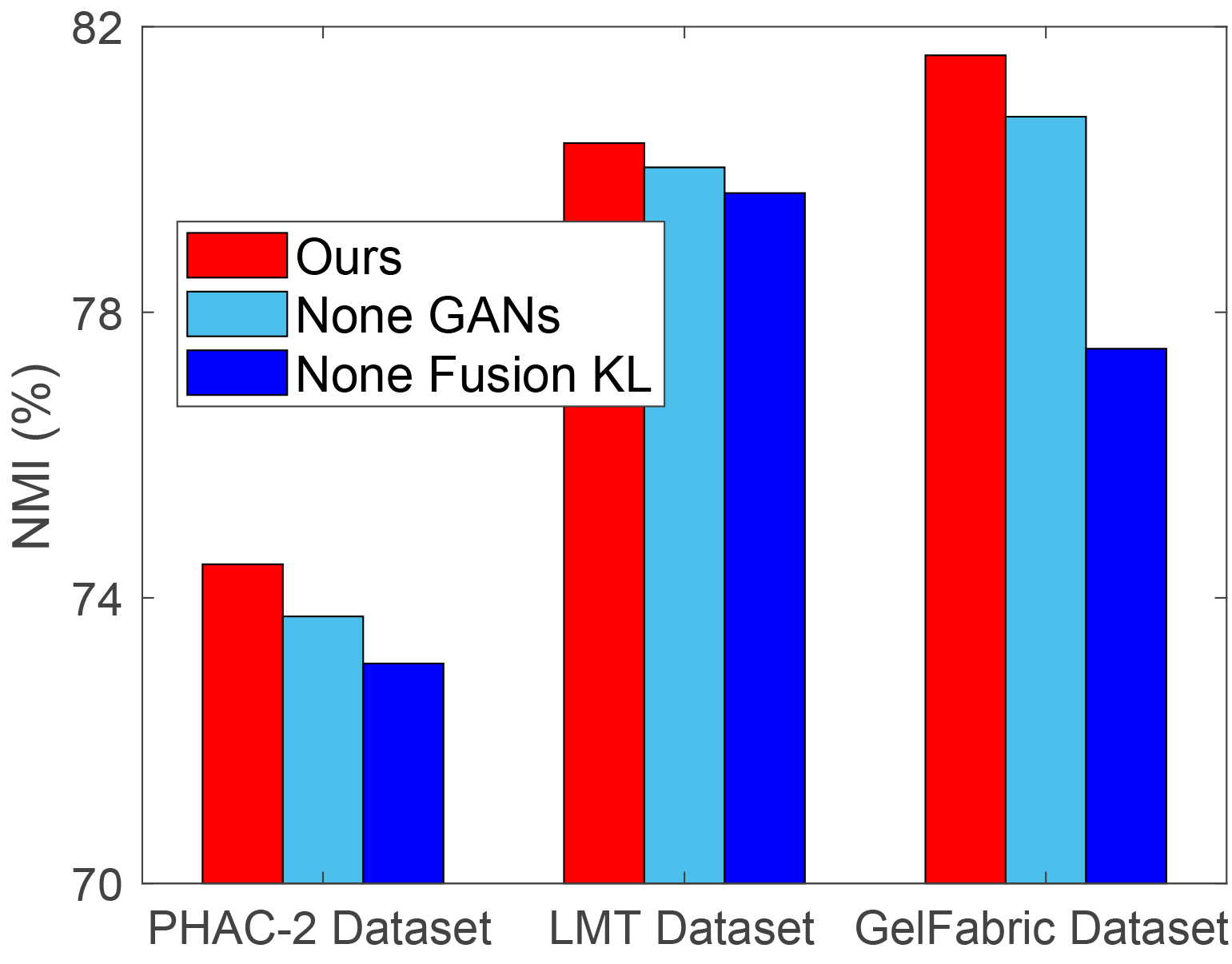}}
	\caption{Effectiveness of 
		cross-modal clustering GANs, encoders, fusion KL-divergence losses, when the missing rate $\mc{MR}$ is 0.1.}
	\label{fig:6}
\end{figure}
For instance, compared with single-modality methods (\emph{i.e.}, \textbf{SC1} and \textbf{SC2}), the performance is raised by $12.68\%$ in ACC and $7.42\%$ in NMI on the PHAC-2 dataset, which demonstrates the fact that fusing visual and tactile modalities does improve the clustering performance. 
The results also show that our model is able to learn complementary information among the heterogeneous data.
Compared with partial multi-view clustering method \textbf{UEAF} and visual-tactile fusing clustering method \textbf{VTFC}, the performance is raised by $1.89\%$ and $3.62\%$ in ACC and NMI, respectively.
The reason why our GPVTF model achieves considerable achievements is that our model can not only complete the missing data but also well align the heterogeneous data.
\textbf{2}) As shown in Figure~\ref{fig:3} and Figure~\ref{fig:4}, our GPVTF model outperforms other methods under different missing rates ($0.1\sim0.5$) on all the three datasets.
Moreover, our model can also achieve competitive results on the PHAC-2 and LMT datasets even though the missing rate is very large.
This observation indicates the effectiveness of the proposed conditional cross-modal clustering GANs.
Besides, although the performance of \textbf{SC2} drops more slowly than ours, its performance is very low in most cases.
We also find an interesting phenomenon that some multi-view clustering methods (\emph{i.e.}, GRMF, IMG and GLMSC) even perform worse than single-view methods.
The possible reason is that these methods do not take the gap between visual and tactile data into account.
Directly fusion the heterogeneous data in a violent way would inevitably lead to performance degradation.

\subsection{Ablation Study}
The effect of the proposed cross-modal clustering GANs, fusion KL-divergence losses are analyzed first. 
Then we report the analyses of most important parameters $\alpha$, $\beta$, $\varphi_1$ and $\varphi_2$.

\textbf{Effectiveness of Cross-Modal Clustering GANs, Fusion KL-Divergence Losses:}
As shown in Figure~\ref{fig:6}, we first conduct ablation study to illustrate the effect of the proposed conditional cross-modal clustering GANs and fusion KL-Divergence losses when missing rate is set to be  0.1, where ``None GANs'' means the proposed conditional cross-modal clustering GANs are not employed and ``None Fusion KL'' means the proposed fusion KL-Divergence losses are not employed, respectively.
We can observe that ``Ours'' outperforms ``None GANs'' among all the datasets, which proves that the proposed conditional cross-modal clustering GANs promotes to achieve better performance.
``Ours'' outperforms than ``None Fusion KL" proves that the proposed fusion KL-Divergence losses could better discover the information hidden in multi-modality data, and further enhance the performance.

\begin{figure}[htbp]
	\centering
	\subcaptionbox{ACC ($\%$) with different $\alpha$}{\includegraphics[width =.49\columnwidth]
		{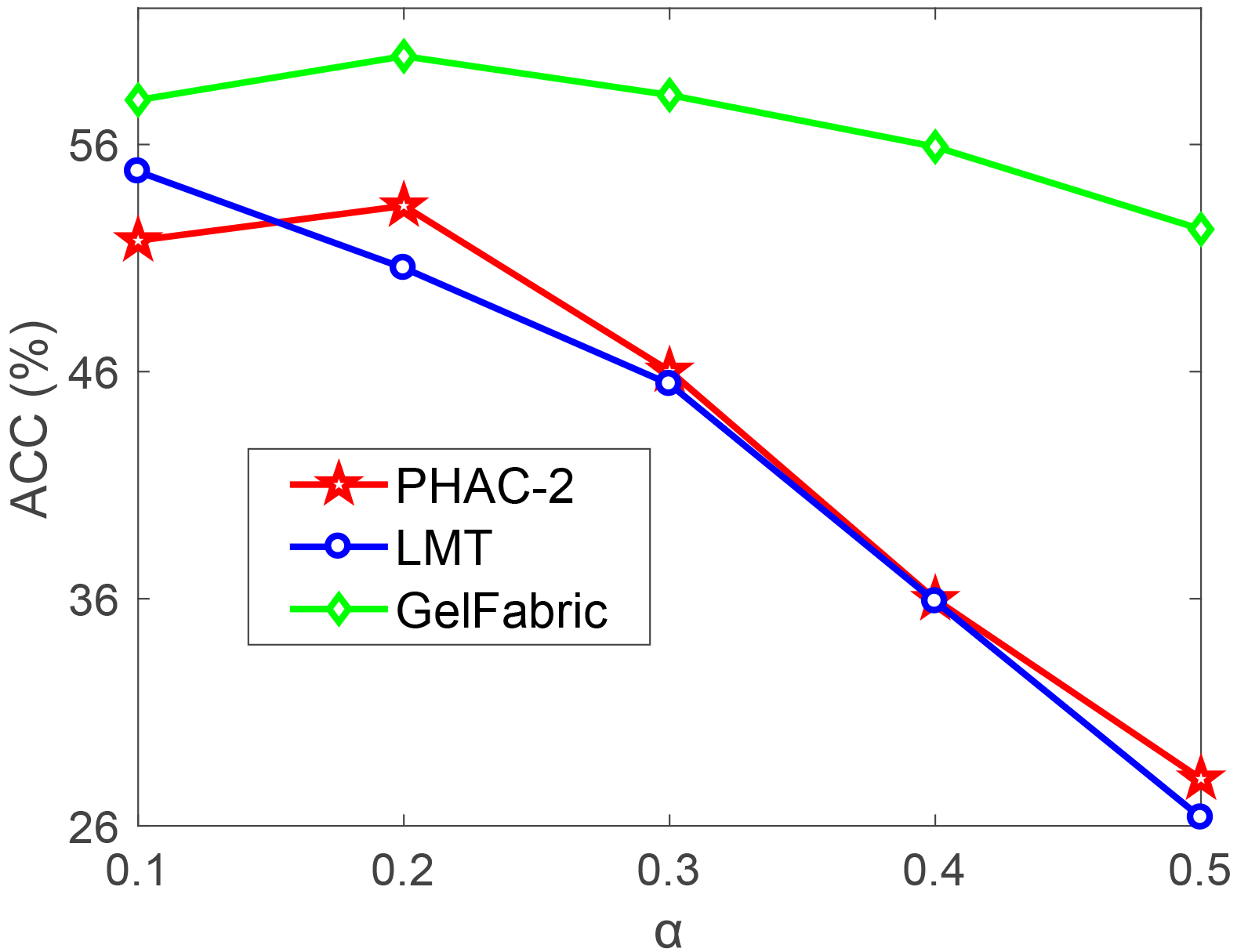}}
	\subcaptionbox{ACC ($\%$) with different $\beta$}{\includegraphics[width =.49\columnwidth]
		{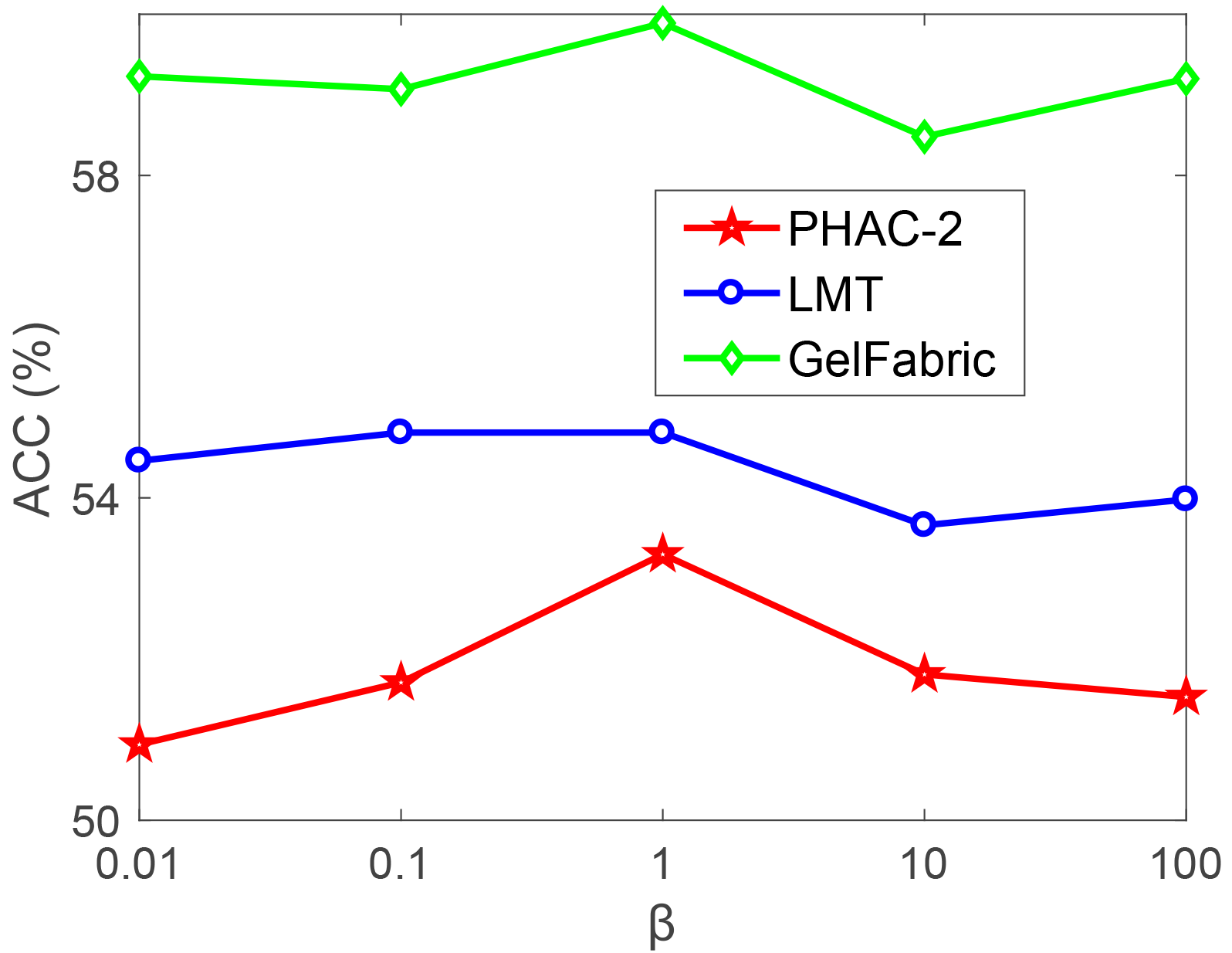}}
	\caption{ACC ($\%$) performance with different $\alpha$ and $\beta$ when the  missing rate $\mc{MR}$ is 0.1.}
	\label{fig:7}
\end{figure}

\begin{figure}[htbp]
	\centering
	\subcaptionbox{NMI ($\%$) performance}{\includegraphics[width =.49\columnwidth]
		{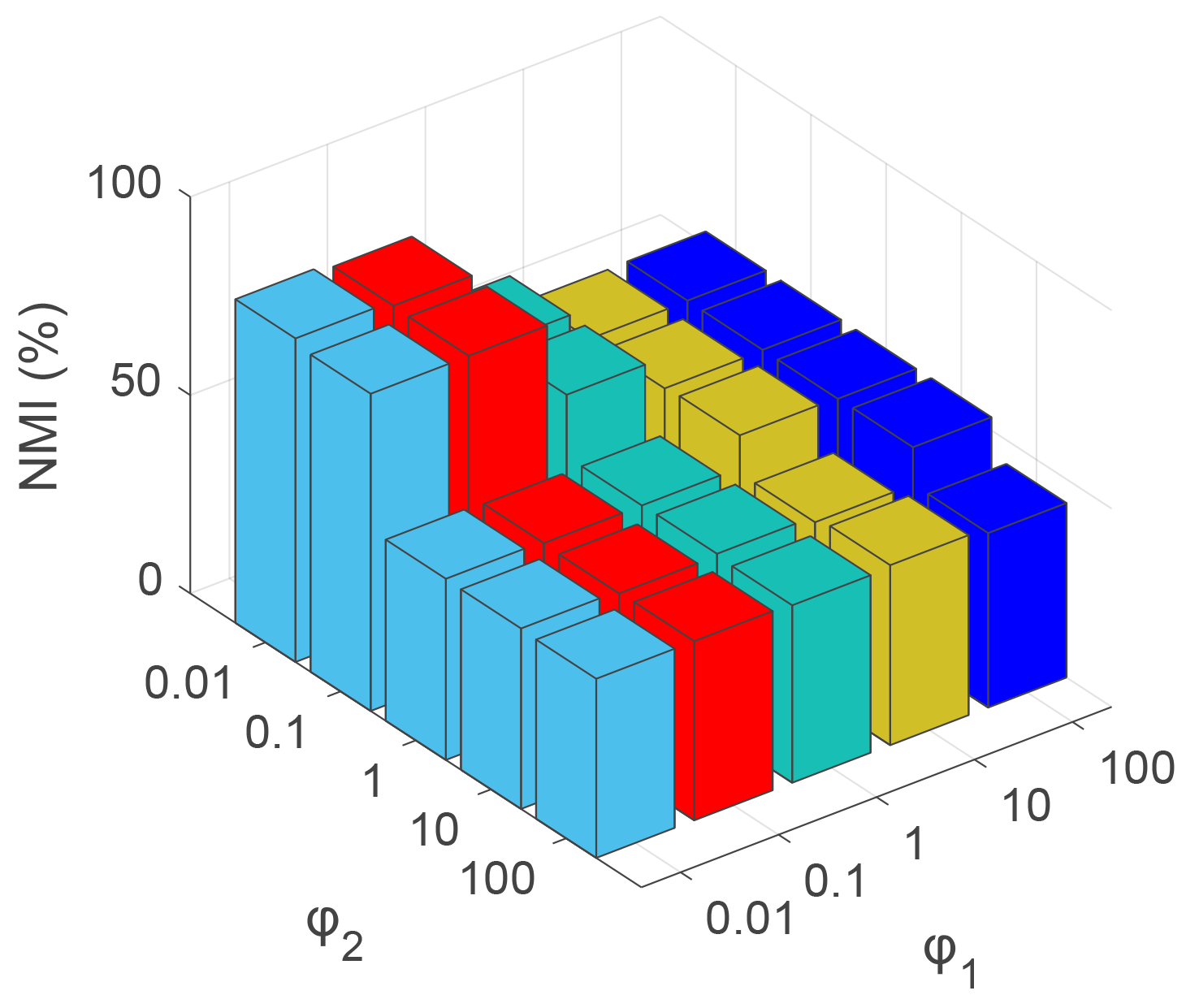}}
	\subcaptionbox{ACC ($\%$) performance}{\includegraphics[width =.49\columnwidth]
		{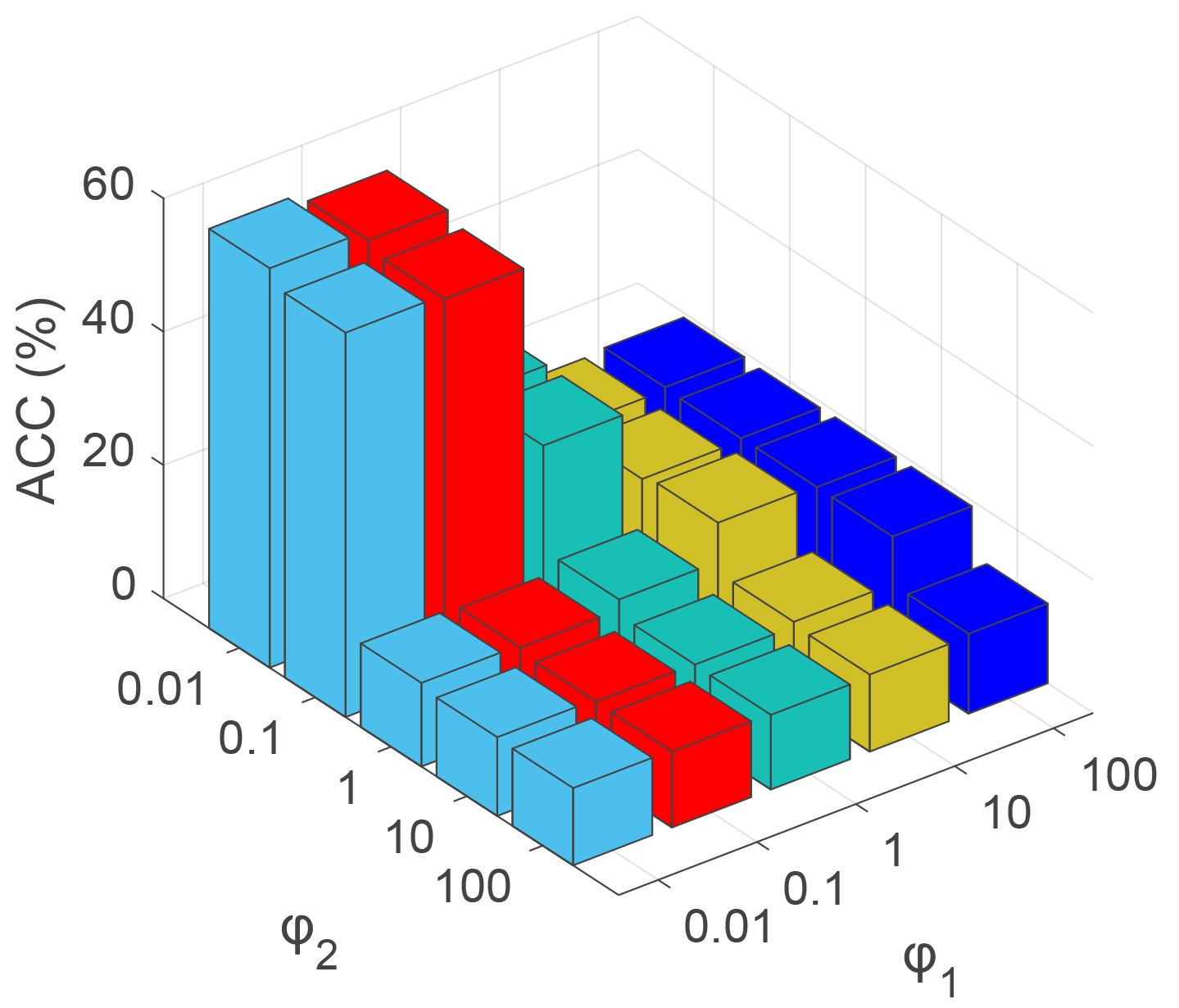}}
	\caption{Performance with different $\varphi_1$ and $\varphi_2$ when the  missing rate $\mc{MR}$ is 0.1.}
	\label{fig:8}
\end{figure}

\textbf{Parameter Analysis:}
To explore the effect important weight coefficient $\alpha$ that controls the proportion of visual and tactile modalities, the parameter $\alpha$ is tuned from the set \{$0.1,0.2,0.3,0.4,0.5$\}, and report the clustering performance in Figure~\ref{fig:7}.
Our model achieves the best clustering results, when the value of $\alpha$ is set to be 0.2, 0.2 and 0.1 on the PHAC-2, GelFabric and LMT datasets, respectively.
Then, the parameter $\beta$ is tuned from the set \{$0.01,0.1,1,10,100$\}, and the ACC performance is plotted in Figure~\ref{fig:7}.
In fact, $\beta$ controls the effect of common component, which further helps to update the encoders $E_1(\cdot)$ and $E_2(\cdot)$ simultaneously. 
It helps to ease the gap between visual and tactile modalities.
It  can be seen that when $\beta$ is set to be $1$, we gain the best performance.
Thus we empirically choose $\beta=1$ as default in this paper. 
To the end, we tune the trade-off parameters $\varphi_1$ and $\varphi_2$ in a similar way with $\beta$. 
As shown in Figure~\ref{fig:8}, our proposed GPVTF model performs best when $\varphi_1$ and $\varphi_2$ are set to be $0.01$.
Thus, we empirically choose $\beta=1$, $\varphi_1=0.01$ and $\varphi_2=0.01$ as default in this paper in order to achieve the best performance.

\section{Conclusion}
In this paper, we put forward a Generative Partial Visual-Tactile Fused (GPVTF) framework, which tries to solve the problem of partial visual-tactile object clustering.
GPVTF completes the partial visual-tactile data via two generators, which generate missing samples conditional on the other modality.
In this way, the performance of clustering can be improved via the completed missing data and the aligned heterogeneous data.
Moreover, pseudo-label based fusion KL-Divergence losses are leveraged to explicitly encapsulate the clustering
task in our network, and further update the modality-specific encoders.
Extensive experimental results on three public real-world benchmark visual-tactile datasets prove the superiority of our framework when comparing with several advanced methods.

\bibliographystyle{aaai21}
\bibliography{mybibfile}

\end{document}